\documentclass[letterpaper, 10 pt, conference]{ieeeconf}  %

\IEEEoverridecommandlockouts                              %

\usepackage{graphics} %
\graphicspath{{figures/}}
\usepackage{amsmath} %
\usepackage{mathtools}
\usepackage{amssymb}  %
\usepackage{siunitx}
\usepackage[noadjust]{cite}
\usepackage{paralist}
\usepackage{xargs}
\usepackage{tikz}
\usepackage{pgfplots}
\usetikzlibrary{angles,patterns,quotes, babel, graphs, pgfplots.groupplots}
\usepackage{tkz-euclide}
\usetkzobj{all}

\definecolor{fzi-darkgreen}{RGB}{0,176,109}
\definecolor{fzi-green}{RGB}{0,108,66}
\definecolor{fzi-lightgreen}{RGB}{115,154,131}
\definecolor{fzi-brightgreen}{RGB}{0,180,111}
\definecolor{fzi-happygreen}{RGB}{146,208,80}
\definecolor{fzi-superlightgreen}{RGB}{191,205,196}
\definecolor{fzi-red}{RGB}{255,0,71}
\definecolor{fzi-blue}{RGB}{0,100,163}
\definecolor{fzi-lightblue}{RGB}{136,195,230}
\definecolor{fzi-yellow}{RGB}{250,187,8}
\usepackage{siunitx}
\makeatletter
\let\NAT@parse\undefined
\makeatother

\usepackage[nolist,nohyperlinks]{acronym}
\usepackage{algorithm}
\usepackage{algpseudocode}
\usepackage{subcaption}
\usepackage{hyperref}

\algnewcommand\algorithmicforeach{\textbf{for each}}

\algdef{S}[FOR]{ForEach}[1]{\algorithmicforeach\ #1\ \algorithmicdo}

\title{\LARGE \bf
Fast Lane-Level Intersection Estimation using\\Markov Chain Monte Carlo Sampling and B-Spline Refinement
}

\author{Annika Meyer$^{1,2}$, Jonas Walter$^{1}$ and Martin Lauer$^{2}$%
\thanks{$^{1}$Annika Meyer and Jonas Walter are with FZI Research Center for Information Technology, Karlsruhe, Germany, {\tt ameyer@fzi.de}}%
\thanks{$^{2}$Annika Meyer and Martin Lauer are with the Institute of Measurement and Control Systems, Karlsruhe Institute of Technology (KIT), Karlsruhe, Germany}%
}

\newcommand\copyrighttext{%

	\footnotesize \textcopyright 2019 IEEE.  Personal use of this material is permitted.  Permission from IEEE must be obtained for all other uses, in any current or future media, including reprinting/republishing this material for advertising or promotional purposes, creating new collective works, for resale or redistribution to servers or lists, or reuse of any copyrighted component of this work in other works.}

\newcommand\copyrightnotice{%

	\begin{tikzpicture}[remember picture,overlay]

	\node[anchor=north,xshift=10pt,yshift=-10pt] at (current page.north) {\fbox{\parbox{\dimexpr\textwidth-\fboxsep-\fboxrule\relax}{\copyrighttext}}};

	\end{tikzpicture}%

}

\begin{document}

\maketitle

\copyrightnotice
\thispagestyle{empty}
\pagestyle{empty}

\begin{acronym}
	\acro{MAP}{maximum a posteriori}
	\acro{MCMC}{Markov chain Monte Carlo}
	\acro{IoU}{Intersection over Union}
\end{acronym}

\begin{abstract}
Estimating the current scene and understanding the potential maneuvers are essential capabilities of automated vehicles. Most approaches rely heavily on the correctness of maps, but neglect the possibility of outdated information.

We present an approach that is able to estimate lanes without relying on any map prior. The estimation is based solely on the trajectories of other traffic participants and is thereby able to incorporate complex environments. In particular, we are able to estimate the scene in the presence of heavy traffic and occlusions.

The algorithm first estimates a coarse lane-level intersection model by Markov chain Monte Carlo sampling and refines it later by aligning the lane course with the measurements using a non-linear least squares formulation. We model the lanes as 1D cubic B-splines and can achieve error rates
of less than 10cm within real-time.

\end{abstract}

\section{Introduction}
Highly accurate maps are seen as essential base for environment perception \cite{kunz_autonomous_2015}.
Maps, that have a high precision and allow for precise localization, ease a lot of difficult perception tasks.
For example, recognizing the road boundary may be simplified to efficient map lookups.

Nevertheless, maps are a static representation of a dynamically changing world.
Road layouts and, thus, the routing are prone to changes due to construction sites leading to expensive and frequent map updates.

Most presented approaches are suitable for lane estimation in simple environments like almost straight roads or highways, but lack the ability to estimate the complexity of urban intersections.
The focus on (simple) lane estimation tasks was mainly motivated by the TuSimple challenge, that provided a highway video dataset with lane boundary annotation.\footnote{The dataset and benchmark was published on benchmark.tusimple.ai, which is no longer available.}. More complex and urban datasets currently lack sufficient and precise lane information in intersections and if anything provide rough estimates on center lines.

\begin{figure}[t]
	\centering
	\includegraphics[width=0.8\linewidth]{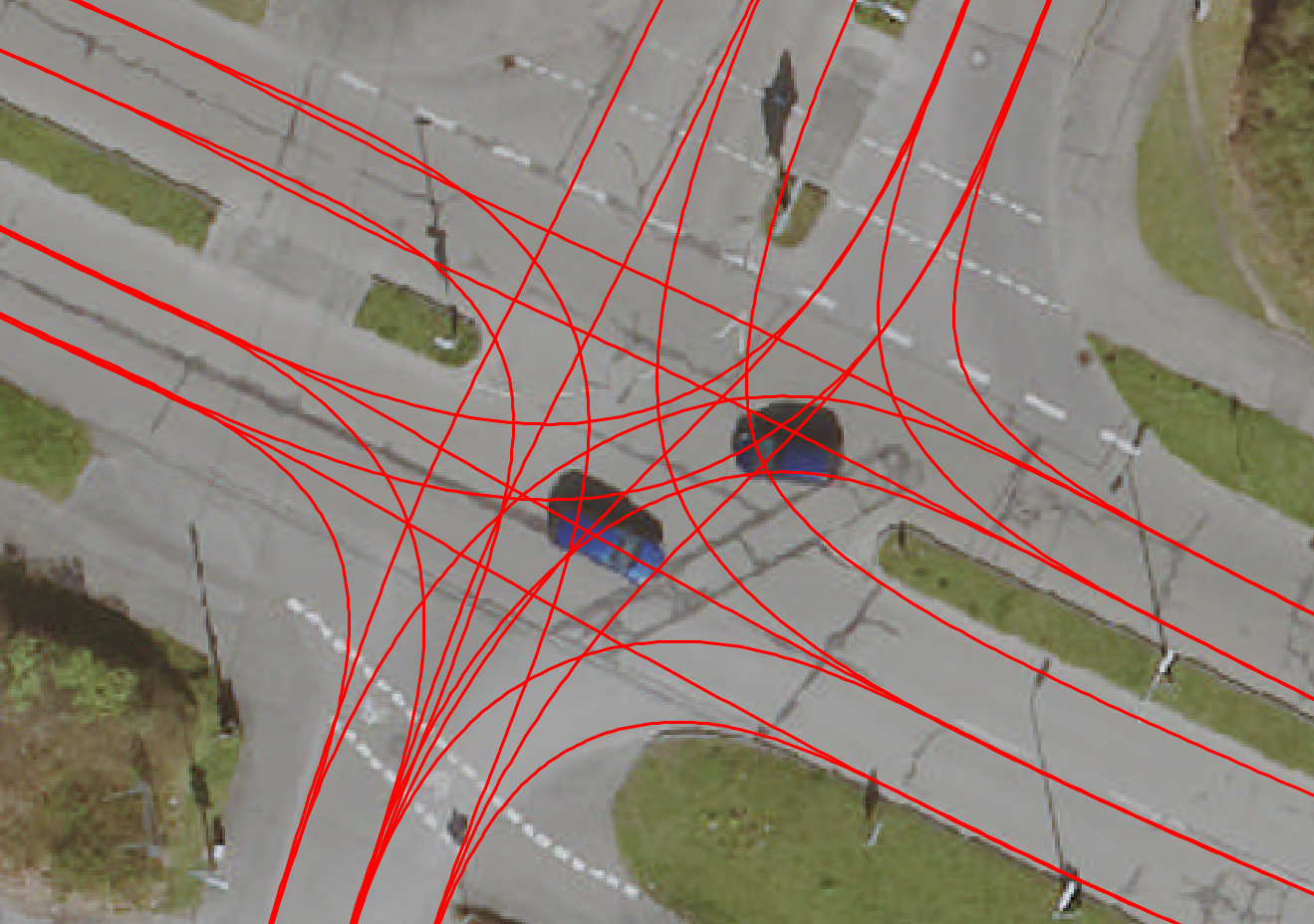}
	\caption{Results of our lane estimation approach. The estimation is based solely on trajectories crossing the intersection. Aerial images: City of Karlsruhe, \texttt{\small www.karlsruhe.de}}
	\label{fig:teaser}
\end{figure}

In the following, we present an approach that is able to estimate the intersection including the lane course precisely as depicted in \autoref{fig:teaser}.
We base the estimation
on the detection and the tracking of other vehicles passing the intersection.

In order to determine the most probable estimate for an intersection at once, the maximum of the posterior distribution has to be calculated, which is highly intractable for the complexity of an intersection model.
\acl{MCMC} is a sampling based approach especially designed to overcome the complexity of the posterior.
Thus, similar to \cite{Meyer_AnytimeLaneLevelIntersection_2019}, we apply a two-stage setting, facilitating \ac{MCMC}.
At first, we estimate the coarse structure of the intersection leading to an initial lane-level model. Here, we examine a multitude of hypotheses on the intersection structure using \ac{MCMC} sampling.
Given the coarse estimate, which we assume to be correct up to the exact lane courses, the lane course estimation can be formulated as a least-squares problem.

We present our work structured as the following:
At first, we examine related work in Section~\ref{sec:related_work}.
In Section~\ref{sec:preproc}, the coarse parameter estimation is presented that estimates the position of arms and the number of lanes as well as the center position of the intersection.
On the basis of this, we present our B-Spline model and the lane course refinement in Section~\ref{sec:model}
We evaluated our approach on simulated trajectories and present the results in Section~\ref{sec:experiments}.

\section{Related Work}
\label{sec:related_work}
In recent years, many approaches dealt with lane estimation and rough intersection estimation.
Only a few approaches aim at precise, lane-level intersection estimation which is necessary for driving across.

Many approaches, for estimating lanes or intersections, used camera images and their cues. However, the image domain is not particularly suitable for huge intersections.
Interesting cues like curbs and markings
might be occluded in heavy traffic and the image coordinate system induces an imprecision for distant areas due to perspective distortion.
This makes it especially hard to detect road areas in huge intersections solely based on camera information, e.g. if a lane or a curbstone is represented by just a few pixels.

Thus, focusing on different sensor data or aggregated camera data is far more promising for more complex environments like urban intersection.
Occlusions could be tackled by e.g. aggregating multiple frames in a topview grid, but still lack the precision in distance.

\paragraph*{Map Generation}
Apart from image cues, some approaches based their estimation on the trajectories of traffic participants. They mostly facilitated fleet data and thus relied on a multitude of measurements.

Roeth et al.~\cite{Roeth_Roadnetworkreconstruction_2016}, for example, determined the connection of incoming and outgoing lanes.
Similar to our approach, they applied a \ac{MCMC} algorithm, but only sample connecting pairs instead of geometric properties or lane courses.

As an extension, they presented an approach that is able to estimate the exact lanes inside of an intersection \cite{Roeth_ExtractingLaneGeometry_2017}. Since they aim at offline map generation, they do not regard any execution time limits and thus rely on complex and time consuming models. Here as well, \ac{MCMC} sampling was successfully applied to overcome the complexity of intersections.

Similarly, Busch et al. \cite{Busch_DiscreteReversibleJump_2019} employed trajectory clustering for estimating the center lines of the lanes.
The cluster representative was modeled with cubic 2D splines and, thus, implicitly smooth.
They also used an \ac{MCMC} based approach, but evaluated different clustering hypotheses without explicit intersection models.

However, approaches aiming at map generation used fleet data and assumed a huge number of trajectory detections per lane and, thus, show increased computation times.
In addition, a majority of these approaches leverage complex models that still can not be applied in a real-time approach with fewer measurements.

\paragraph*{Online Lane Estimation}
On comparison, Geiger et al. \cite{geiger_3d_2014} estimated the intersection geometry in an online manner using the RGB images of an ego vehicle.
The approach achieved promising results because they incorporated object detections instead of the direct use of camera data. The viewing angle in the dataset, however, was limited to a single camera and lacked sensors facing to the sides. This was sufficient for the specific dataset with only small intersections.
Consistent with this, the intersection model was limited and assumed
only a single lane per driving direction
and collinearity of the crossing arms.
This might have been sufficient, but will not suffice for big, urban intersections.

In 2019, we~\cite{Meyer_AnytimeLaneLevelIntersection_2019} presented a first approach on estimating the lane course of intersections within real-time. Here, we used the trajectories of other traffic participants as measurements, but only required a few of them.
We estimated the lanes in a two-stage approach fully based on \ac{MCMC} sampling.
First, we estimated the intersection parameters and, based on that, we refined the lanes using a multitude of samples similarly in the manner of \ac{MCMC}. We showed that the division into two steps reduces the search space and, thus, the computation time radically.
We could estimate the coarse parameters in less than \SI{50}{\milli\second}, but experienced slow convergence for the lane estimation with execution times above \SI{100}{\milli\second}. In addition the resulting lane courses lacked smoothness despite an additional smoothness term in the \ac{MCMC} evaluation.

Thus, in this work we present how to overcome the drawbacks of a purely \ac{MCMC} based approach and emphasize the advantages of representing lanes with B-splines.

\section{Intersection Parameter Estimation}
\label{sec:preproc}

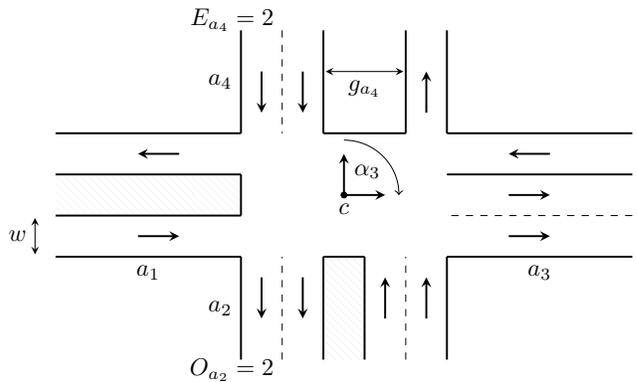
\begin{figure}[tb]
	\begin{center}
		\resizebox{\linewidth}{!}{

  \begin{tikzpicture}[
      scale=0.6,
      arrow/.style={thin,<->,shorten >=1pt,shorten <=1pt,>=stealth},
      larrow/.style={thick,<-,>=stealth},
      rarrow/.style={thick,->,>=stealth}
    ]
\draw [thick] (-7,-1.5) -- (-2.5,-1.5)node[midway,below]{$a_1$};
\draw [thick] (-7,1.5) -- (-2.5,1.5);
\draw [thick] (-7,0.5) -- (-2.5,0.5);
\draw [thick] (-7,-0.5) -- (-2.5,-0.5);
\draw [thick] (-2.5,0.5) -- (-2.5,-0.5);

\draw [thick] (7,-1.5) -- (2.5,-1.5)node[midway,below]{$a_3$};
\draw [thick] (7,1.5) -- (2.5,1.5) node (origo) {};
\draw [thick] (7,0.5) -- (2.5,0.5);
\draw [thin, dashed] (7,-0.5) -- (2.5,-0.5);

\draw [thick] (-2.5,-4) -- (-2.5,-1.5)node[align=center,midway,left]{$a_2$};
\draw [thin,dashed] (-1.5,-4) -- (-1.5,-1.5)  node [pos=-0.1,left]{$O_{a_2}=2$};
\draw [thick] (0.5,-4) -- (0.5,-1.5);
\draw [thick] (-0.5,-1.5) -- (0.5,-1.5);
\draw [thick] (-0.5,-4) -- (-0.5,-1.5);
\draw [thin, dashed] (1.5,-4) -- (1.5,-1.5);
\draw [thick] (2.5,-4) -- (2.5,-1.5);

\draw [thick] (-2.5,4) -- (-2.5,1.5)node[midway,left]{$a_4$};
\draw [thin,dashed] (-1.5,4) -- (-1.5,1.5)node[pos=-0.1,left]{$E_{a_4}=2$};
\draw [thick] (-0.5,1.5) -- (1.5,1.5);
\draw [thick] (-0.5,4) -- (-0.5,1.5);
\draw [thick] (1.5,4) -- (1.5,1.5);
\draw [thick] (2.5,4) -- (2.5,1.5)node (beta) {};

\tkzDefPoint(0,1){C}
\tkzDefPoint(0,0){B}
\tkzDefPoint(1,0){A}
\pic [draw, <-, angle radius=8mm, angle eccentricity=0.6, "$\alpha_{3}$"] {angle = A--B--C};

\draw[opacity=0.2, draw=none, pattern=north west lines, pattern color=black] (-7,-0.5) rectangle (-2.5,0.5);
\draw[opacity=0.2, draw=none, pattern=north west lines, pattern color=black] (-0.5, -4) rectangle (0.5, -1.5);
\draw[opacity=0.1, draw=none, pattern=north west lines, pattern color=black] (-0.5, 4) rectangle (1.5, 1.5);

\draw [arrow] (-0.5,3) -- (1.5,3)node[midway,below]{$g_{a_4}$};
\draw [arrow] (-7.5,-0.5) -- (-7.5,-1.5)node[align=center,midway,left]{$w$};

\draw [rarrow] (-5,-1) -- (-4,-1);\draw [larrow] (-5,1) -- (-4,1);
\draw [larrow] (5,-1) -- (4,-1);\draw [rarrow] (5,1) -- (4,1);
\draw [larrow] (5,0) -- (4,0);

\draw [larrow] (2,3) -- (2,2);
\draw [rarrow] (-1,3) -- (-1,2);\draw [rarrow] (-2,3) -- (-2,2);
\draw [rarrow] (1,-3) -- (1,-2);\draw [rarrow] (2,-3) -- (2,-2);
\draw [larrow] (-1,-3) -- (-1,-2);\draw [larrow] (-2,-3) -- (-2,-2);

\draw [rarrow] (0,0) -- (0,1);
\draw [rarrow] (0,0) -- (1,0);
\filldraw (0,0) circle (2pt) node[align=right,below] {$c$};

\end{tikzpicture}
 		}
	\end{center}
	\caption{Parametric model of an intersection $I=(c,A)$. The base is a center point $c$ with a set of arms $A$. Each arm is defined as $a = (\alpha, g, E, O, w)$ with angle $\alpha$ describing the orientation in the intersection and $g$ enabling a structural separation between driving directions on the same arm. The number of entries $E$ and exits $O$ in conjunction with the width $w$ of the lanes determine the overall layout of the intersection. }
	\label{fig:topo_model}
\end{figure}

\begin{figure*}[t]
	\centering
	\begin{subfigure}[b]{0.321\linewidth}
		\centering
		\resizebox{\linewidth}{!}{

\begin{tikzpicture}[
scale=0.6,
arrow/.style={thin,<->,shorten >=1pt,shorten <=1pt,>=stealth},
larrow/.style={thick,<-,>=stealth},
rarrow/.style={thick,->,>=stealth}
]

\draw [thick] (-7,-1.5) -- (-2.5,-1.5);
\draw [thick] (-7,1.5) -- (-2.5,1.5);
\draw [thick] (-7,0.5) -- (-2.5,0.5);
\draw [thick] (-7,-0.5) -- (-2.5,-0.5);
\draw [thick] (-2.5,0.5) -- (-2.5,-0.5);

\draw [thick] (7,-1.5) -- (2.5,-1.5);
\draw [thick] (7,1.5) -- (2.5,1.5) node (origo) {};
\draw [thick] (7,0.5) -- (2.5,0.5);
\draw [thin, dashed] (7,-0.5) -- (2.5,-0.5);

\draw [thick] (-2.5,-4) -- (-2.5,-1.5);
\draw [thin,dashed] (-1.5,-4) -- (-1.5,-1.5);
\draw [thick] (0.5,-4) -- (0.5,-1.5);
\draw [thick] (-0.5,-1.5) -- (0.5,-1.5);
\draw [thick] (-0.5,-4) -- (-0.5,-1.5);
\draw [thin, dashed] (1.5,-4) -- (1.5,-1.5);
\draw [thick] (2.5,-4) -- (2.5,-1.5);

\draw [thick] (-2.5,4) -- (-2.5,1.5);
\draw [thin,dashed] (-1.5,4) -- (-1.5,1.5);
\draw [thick] (-0.5,1.5) -- (0.5,1.5);
\draw [thick] (-0.5,4) -- (-0.5,1.5);
\draw [thick] (1.5,4) -- (1.5,1.5);
\draw [thick] (0.5,4) -- (0.5,1.5);
\draw [thick] (2.5,4) -- (2.5,1.5)node (beta) {};

\draw[opacity=0.2, draw=none, pattern=north west lines, pattern color=black] (-7,-0.5) rectangle (-2.5,0.5);
\draw[opacity=0.2, draw=none, pattern=north west lines, pattern color=black] (-0.5, -4) rectangle (0.5, -1.5);
\draw[opacity=0.1, draw=none, pattern=north west lines, pattern color=black] (-0.5, 4) rectangle (0.5, 1.5);

\draw [rarrow] (-5,-1) -- (-4,-1);\draw [larrow] (-5,1) -- (-4,1);
\draw [larrow] (5,-1) -- (4,-1);\draw [rarrow] (5,1) -- (4,1);
\draw [larrow] (5,0) -- (4,0);

\draw [larrow] (2,3) -- (2,2);
\draw [larrow] (1,3) -- (1,2);
\draw [rarrow] (-1,3) -- (-1,2);\draw [rarrow] (-2,3) -- (-2,2);
\draw [rarrow] (1,-3) -- (1,-2);\draw [rarrow] (2,-3) -- (2,-2);
\draw [larrow] (-1,-3) -- (-1,-2);\draw [larrow] (-2,-3) -- (-2,-2);
\filldraw (0,0) circle (2pt);

\draw [dash pattern=on \pgflinewidth off 5pt, line width=3pt, line cap=round, fzi-yellow] (2,4) -- (2, 1.5) -- (2.15, 1.15) -- (2.5, 1) -- (7, 1);
\draw [dash pattern=on \pgflinewidth off 5pt, line width=3pt, line cap=round, fzi-lightblue] (7,-1) -- (-7, -1);
\draw [dash pattern=on \pgflinewidth off 5pt, line width=3pt, line cap=round, fzi-red] (2,-4) -- (2, 4);
\draw [dash pattern=on \pgflinewidth off 5pt, line width=3pt, line cap=round, fzi-happygreen] (-1,4) -- (-1, -4);

\node[inner sep=0pt] (green) at (-1.1,2.2) {\includegraphics[width=0.12\textwidth]{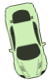}};
\node[inner sep=0pt] (red) at (2,-2.5) {\includegraphics[width=0.14\textwidth]{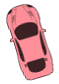}};
\node[inner sep=0pt] (yellow) at (4,1) {\includegraphics[width=0.18\textwidth]{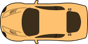}};

\end{tikzpicture}
 		}
		\caption{Coarse Parameter Estimation}
		\label{fig:topo_traj}
	\end{subfigure}
	\begin{subfigure}[b]{0.321\linewidth}
		\centering
		\resizebox{\linewidth}{!}{

\begin{tikzpicture}[
scale=0.6,
arrow/.style={thin,<->,shorten >=1pt,shorten <=1pt,>=stealth},
larrow/.style={thick,<-,>=stealth},
rarrow/.style={thick,->,>=stealth}
]
\draw [thick] (-7,-1.5) -- (-2.5,-1.5);
\draw [thick] (-7,1.5) -- (-2.5,1.5);
\draw [thick] (-7,0.5) -- (-2.5,0.5);
\draw [thick] (-7,-0.5) -- (-2.5,-0.5);
\draw [thick] (-2.5,0.5) -- (-2.5,-0.5);

\draw [thick] (7,-1.5) -- (2.5,-1.5);
\draw [thick] (7,1.5) -- (2.5,1.5) node (origo) {};
\draw [thick] (7,0.5) -- (2.5,0.5);
\draw [thin, dashed] (7,-0.5) -- (2.5,-0.5);

\draw [thick] (-2.5,-4) -- (-2.5,-1.5);
\draw [thin,dashed] (-1.5,-4) -- (-1.5,-1.5);
\draw [thick] (0.5,-4) -- (0.5,-1.5);
\draw [thick] (-0.5,-1.5) -- (0.5,-1.5);
\draw [thick] (-0.5,-4) -- (-0.5,-1.5);
\draw [thin, dashed] (1.5,-4) -- (1.5,-1.5);
\draw [thick] (2.5,-4) -- (2.5,-1.5);

\draw [thick] (-2.5,4) -- (-2.5,1.5);
\draw [thin,dashed] (-1.5,4) -- (-1.5,1.5);
\draw [thick] (-0.5,1.5) -- (0.5,1.5);
\draw [thick] (-0.5,4) -- (-0.5,1.5);
\draw [thick] (1.5,4) -- (1.5,1.5);
\draw [thick] (0.5,4) -- (0.5,1.5);
\draw [thick] (2.5,4) -- (2.5,1.5)node (beta) {};

\fill[fzi-lightblue, nearly transparent]  (-7,-1.5) -- (-7,-0.5) -- (-0.3,-0.5) -- (-0.3,-1.5) -- cycle;
\draw[fzi-lightblue, thick]  (-7,-1.5) -- (-7,-0.5) -- (-0.3,-0.5) -- (-0.3,-1.5) -- cycle;
\fill[fzi-lightblue, nearly transparent]  (0.3,-1.5) -- (0.3,-0.5) -- (7,-0.5) -- (7,-1.5) -- cycle;
\draw[fzi-lightblue, thick]  (0.3,-1.5) -- (0.3,-0.5) -- (7,-0.5) -- (7,-1.5) -- cycle;
\node  at (-5, -2) {$e_1$};
\node  at (5, -2) {$o_1$};

\fill[fzi-red, nearly transparent]  (1.5,-4) -- (1.5,-0.3) -- (2.5,-0.3) -- (2.5,-4) -- cycle;
\draw[fzi-red, thick]  (1.5,4) -- (1.5,0.3) -- (2.5,0.3) -- (2.5,4) -- cycle;
\fill[fzi-red, nearly transparent]  (1.5,4) -- (1.5,0.3) -- (2.5,0.3) -- (2.5,4) -- cycle;
\draw[fzi-red, thick]  (1.5,-4) -- (1.5,-0.3) -- (2.5,-0.3) -- (2.5,-4) -- cycle;
\node  at (3, -3) {$e_2$};

\fill[fzi-happygreen, nearly transparent]  (-0.5,-4) -- (-0.5,-0.3) -- (-1.5,-0.3) -- (-1.5,-4) -- cycle;
\draw[fzi-happygreen, thick]  (-0.5,-4) -- (-0.5,-0.3) -- (-1.5,-0.3) -- (-1.5,-4) -- cycle;
\fill[fzi-happygreen, nearly transparent]  (-0.5,0.3) -- (-0.5,4) -- (-1.5,4) -- (-1.5,0.3) -- cycle;
\draw[fzi-happygreen, thick]  (-0.5,0.3) -- (-0.5,4) -- (-1.5,4) -- (-1.5,0.3) -- cycle;
\node  at (-0.1, -3) {$o_4$};
\node  at (-0.1, 3) {$e_4$};

\fill[fzi-yellow, nearly transparent] (2.8, 0.5) -- (7, 0.5) -- (7,1.5) -- (2.8,1.5) -- cycle;
\draw[fzi-yellow, thick] (2.8, 0.5) -- (7, 0.5) -- (7,1.5) -- (2.8,1.5) -- cycle;
\fill[fzi-yellow, nearly transparent] (1.5,4) -- (1.5, 1.5) -- (2.5, 1.5) -- (2.5,4) --  cycle;
\draw[fzi-yellow, dashed, thick] (1.5,4) -- (1.5, 1.5) -- (2.5, 1.5) -- (2.5,4) --  cycle;
\node  at (5, 2) {$e_3$};
\node  at (3.5, 3) {$o_2/o_3$};

\draw [dash pattern=on \pgflinewidth off 5pt, line width=3pt, line cap=round, fzi-yellow] (2,4) -- (2, 1.5) -- (2.15, 1.15) -- (2.5, 1) -- (7, 1);
\draw [dash pattern=on \pgflinewidth off 5pt, line width=3pt, line cap=round, fzi-lightblue] (7,-1) -- (-7, -1);
\draw [dash pattern=on \pgflinewidth off 5pt, line width=3pt, line cap=round, fzi-red] (2,-4) -- (2, 4);
\draw [dash pattern=on \pgflinewidth off 5pt, line width=3pt, line cap=round, fzi-happygreen] (-1,4) -- (-1, -4);

\draw[opacity=0.2, draw=none, pattern=north west lines, pattern color=black] (-7,-0.5) rectangle (-2.5,0.5);
\draw[opacity=0.2, draw=none, pattern=north west lines, pattern color=black] (-0.5, -4) rectangle (0.5, -1.5);
\draw[opacity=0.1, draw=none, pattern=north west lines, pattern color=black] (-0.5, 4) rectangle (0.5, 1.5);

\draw [rarrow] (-5,-1) -- (-4,-1);\draw [larrow] (-5,1) -- (-4,1);
\draw [larrow] (5,-1) -- (4,-1);\draw [rarrow] (5,1) -- (4,1);
\draw [larrow] (5,0) -- (4,0);

\draw [larrow] (2,3) -- (2,2);
\draw [larrow] (1,3) -- (1,2);
\draw [rarrow] (-1,3) -- (-1,2);\draw [rarrow] (-2,3) -- (-2,2);
\draw [rarrow] (1,-3) -- (1,-2);\draw [rarrow] (2,-3) -- (2,-2);
\draw [larrow] (-1,-3) -- (-1,-2);\draw [larrow] (-2,-3) -- (-2,-2);
\filldraw (0,0) circle (2pt);

\end{tikzpicture}
 		}
		\caption{Lane Stubs \& Initialization}
		\label{fig:split}
	\end{subfigure}
	\begin{subfigure}[b]{0.321\linewidth}
		\centering
		\resizebox{\linewidth}{!}{

    \begin{tikzpicture}[
      scale=0.6,
      arrow/.style={thin,<->,shorten >=1pt,shorten <=1pt,>=stealth},
      larrow/.style={thick,<-,>=stealth},
      rarrow/.style={thick,->,>=stealth}
    ]
\foreach \y in {1.5,0.5,-0.5,-1.5}{
	\draw [thick] plot [smooth] coordinates { (-7,\y) (-5,\y) (-3,\y) (-1,\y) (1,\y) (3,\y) (5,\y) (7,\y)};
}

\foreach \sign in {-1,1}{
	\foreach \x in {0.5,1.5,2.5}{
		\draw [thick] plot [smooth] coordinates {
			  (\sign*\x,\sign*4)
		      (\sign*\x+\sign*0.1*1*1,\sign*4-\sign*1.88*1)
		      (\sign*\x+\sign*0.1*2*2,\sign*4-\sign*1.88*2)
		      (\sign*\x+\sign*0.1*3*3,\sign*4-\sign*1.88*3)
			  (\sign*\x+\sign*0.1*4*4,\sign*4-\sign*1.88*4)
		};
	}
}

\draw [fzi-red, dash pattern=on \pgflinewidth off 5pt, line width=3pt, line cap=round, ] plot [smooth] coordinates {
	(0.5+1.5,1*4)
	(0.5+1.5+1*0.1*1*1,1*4-1*1.88*1)
	(0.5+1.5+1*0.1*2*2,1*4-1*1.88*2)
	(0.5+1.5+1*0.1*3*3,1*4-1*1.88*3)
	(0.5+1.5+1*0.1*4*4,1*4-1*1.88*4)
};

\draw [fzi-happygreen, dash pattern=on \pgflinewidth off 5pt, line width=3pt, line cap=round] plot [smooth] coordinates {
	(0.5+-1.5,-1*4)
	(0.5+-1.5+-1*0.1*1*1,-1*4--1*1.88*1)
	(0.5+-1.5+-1*0.1*2*2,-1*4--1*1.88*2)
	(0.5+-1.5+-1*0.1*3*3,-1*4--1*1.88*3)
	(0.5+-1.5+-1*0.1*4*4,-1*4--1*1.88*4)
};

\draw [fzi-yellow, dash pattern=on \pgflinewidth off 5pt, line width=3pt, line cap=round] plot [smooth] coordinates { (6.5,1) (4.5,1.05) (2.65,1.65) (2,3.5) (1.9,4) };

\draw [fzi-lightblue, dash pattern=on \pgflinewidth off 5pt, line width=3pt, line cap=round, ] plot [smooth] coordinates {(-7,-1) (-5,-1) (-3,-1) (-1,-1) (1,-1) (3,-1) (5,-1) (7,-1)};

\draw [thick] plot [smooth] coordinates { (6.5,1.5) (4.5,1.55) (3,2) (2.55,3.5) (2.5,4)};
\draw [thick] plot [smooth] coordinates {(6.5,0.5) (4.5,0.55) (2.35,1.35) (1.55,3.5) (1.5,4)};

\draw[opacity=0.2, draw=none, pattern=north west lines, pattern color=black] (-7,-0.5) rectangle (-3,0.5);

\draw [rarrow] (-5,-1) -- (-4,-1);\draw [larrow] (-5,1) -- (-4,1);
\draw [larrow] (5,-1) -- (4,-1);\draw [rarrow] (5,1) -- (4,1);
\draw [larrow] (5,0) -- (4,0);

\draw [larrow] (1,3) -- (1,2);\draw [larrow] (2,3) -- (2,2);
\draw [rarrow] (-2.3,3) -- (-2,2);\draw [rarrow] (-3.3,3) -- (-3,2);
\draw [rarrow] (2.3,-3) -- (2,-2);\draw [rarrow] (3.3,-3) -- (3,-2);
\draw [larrow] (-1,-3) -- (-1,-2);\draw [larrow] (-2,-3) -- (-2,-2);

\node[inner sep=0pt] (green) at (-2.1,2.2) {\includegraphics[width=0.12\textwidth]{figures/tikz/car_fzi_green.png}};
\node[inner sep=0pt] (red) at (3.2,-2.5) {\includegraphics[width=0.14\textwidth]{figures/tikz/car_fzi_red.png}};
\node[inner sep=0pt] (yellow) at (5,1) {\includegraphics[width=0.18\textwidth]{figures/tikz/car_fzi_yelloq.png}};

    \end{tikzpicture}
 		}
		\caption{Lane Course Refinement}
		\label{fig:lane_traj}
	\end{subfigure}
	\caption{Two-steps approach: Using only trajectories of other traffic participants, we first estimate a coarse parameter model (a) using \acs{MCMC} sampling. For this, we split the trajectories into two distinct parts at the closest point to the estimated center (b). The estimated parts are connected using the trajectories and used as initial lane courses for a novel B-spline refinement (c).}
	\label{fig:triple_teaser}
\end{figure*}

Analogous to \cite{Meyer_AnytimeLaneLevelIntersection_2019, Roeth_Roadnetworkreconstruction_2016, geiger_3d_2014, Roeth_ExtractingLaneGeometry_2017, Busch_DiscreteReversibleJump_2019}, we apply \ac{MCMC} sampling to the problem of estimating the coarse intersection parameters. Thus, we sample different intersection models from a Markov chain, accepting a new estimate only, if it is sufficiently close or better than the previous estimate.

\subsection{Intersection Parameter Model}

We use the model presented in \cite{Meyer_AnytimeLaneLevelIntersection_2019}, that is constructed as depicted in \autoref{fig:topo_model}.
An intersection $I=(c,A)$ is based at a center point $c$. It is comprised of a set of arms $A$ with arm $a = (\alpha, g, E, O, w)$ having several parameters assigned. An arm $a$ has an angle $\alpha$ describing the orientation in the intersection according to its coordinate system. $g$ enables a structural separation between driving directions on the same arm and describes the gap between the lanes with different driving directions. The number of intersection entries $E$ and exits $O$ in conjunction with the width $w$ of the lanes determine the overall layout of an intersection.

\subsection{\acl{MCMC} Sampling}
\label{subsec:mcmc}

We start with an initial intersection model $I$ and iteratively modify a single parameter according to Algorithm \autoref{al:topo}. The step probabilities presented here, have been empirically determined and might differ for different driving scenarios.

Adding an arm to the model, requires a more sophisticated approach than the other sampling steps.
A new arm is sampled by drawing from a uniform distribution spread across the gaps between existing arms while maintaining a minimum angular distance of \SI{25}{\degree} to the existing arms.

Most of the sampling steps follow the detailed balance equation required by the Metropolis algorithm~\cite{metropolis_mh_1953}. However, when modifying the number of arms of the estimate, we did not design a symmetric step.
However, the Metropolis-Hastings algorithm~\cite{metropolis_mh_1953} allows for incorporating imbalanced sampling probabilities.
In addition, we extend (and reduce) the state space, when adding (and removing) and arm, so we need to model reversible jumps as well.

Thus, in each iteration, we evaluate the posterior probability $P(I'|Z)$ and perform a Metropolis-Hastings step~\cite{metropolis_mh_1953}, where the trajectory data $Z$ is evaluated against the new model $I'$.

In summary, all models $I'$ that can not comply with the acceptance criterion\footnote{The Jacobian for the reversible jump is neglected here.}
\begin{equation}
\begin{split}
u \leq A(I',I) = \frac{P(I'|Z) P(I \rightarrow I')}{P(I|Z) P(I' \rightarrow I)}  ^{\frac{1}{T_s}},
\label{eq:mh}
\end{split}
\end{equation}
are rejected from the sampling and future models are only based on the previous model $I$. Here, $u \leftarrow \mathcal{U}[0,1]$ is a uniformly distributed random variable. For better convergence we apply simulated annealing with $T_s$ as the annealing parameter.

By following these principles we will only accept samples, that follow the target posterior distribution of $P(I|Z)$. Thus, the advantage of \ac{MCMC} is that we only need to sample from a simple prior distribution $P(I)$ and thereby estimate the maximum of the more complex posterior distribution $P(I|Z)$. This is especially necessary, if the posterior is intractable like here.

\begin{algorithm}[tb]
	\begin{algorithmic}[0]
		\State $\omega \leftarrow \mathcal{U}[0,1]$
		\If {$\omega < 0.3$}

		{rotate a random arm by $\Delta \alpha \leftarrow \mathcal{U}[\ang{-6},\ang{6}]$}
		\ElsIf {$\omega < 0.52$}	\par
		shift center by $\{\Delta c,\phi\} \leftarrow \mathcal{U}([\SI{0}{\meter},\SI{6}{\meter}]\times[0,2\pi])$
		\ElsIf {$\omega < 0.68$}\par
		change gap by $\Delta g \leftarrow \mathcal{U}[\SI{-1.8}{\meter},\SI{1.8}{\meter}]$
		\ElsIf {$\omega < 0.73$}\par
		\State $\theta \leftarrow \mathcal{U}[0,1]$
		\If {$\theta < 0.5$} add arm (see \autoref{subsec:mcmc})
		\Else \hspace{0.1em} remove arm $a \leftarrow \mathcal{U}_D(A)$
		\EndIf
		\Else
		\State $\theta \leftarrow \mathcal{U}[0,1]$
		\If {$\theta < 0.5$}\par
		\State $\gamma \leftarrow \mathcal{U}[0,1]$
		\If {$\gamma < 0.5$} add lane to the left
		\Else \hspace{0.1em} add lane to the right
		\EndIf
		\Else \hspace{0.1em} remove lane $l \leftarrow \mathcal{U}(E \lor O)$ %
		\EndIf
		\EndIf
	\end{algorithmic}
	\caption{Sample new intersections by modifying a single parameter at a step.}
	\label{al:topo}
\end{algorithm}

Please note, that we do not estimate the width of the lanes, but assume a fixed value $w=$~\SI{2.7}{\meter}. Since we base our estimation on trajectories and, thus, use cues related to the centerline of a lane, the width is hardly observable.

The evaluation of the model requires the calculation of the posterior probability $P(I|Z)$. Because we only need a relation between the posterior probabilities of two intersections as shown in \autoref{eq:mh} and according to the rules of Bayesian theory we only need to calculate the likelihood $P(Z|I)$ and a prior probability $P(I)$ for each intersection. Thus the acceptance criterion becomes
\begin{equation}
\begin{split}
u \leq A(I',I) = \frac{P(I') P(Z|I') P(I \rightarrow I')}{P(I) P(Z|I) P(I' \rightarrow I)}  ^{\frac{1}{T_s}}.
\label{eq:mh_fraction}
\end{split}
\end{equation}
The prior $P(I)$ is modeled uninformative.

Assuming independent measurements
\begin{equation}
P(Z|I) = \prod^Z_z P(z|I).
\end{equation}
For the likelihood $P(z|I)$ of each measurement $z = (x,y,\phi,t,id)$, we further assume the positions $(x,y)$ of the trajectories to be normally distributed around the center line of the lane. The angular deviation of the orientation $\phi$ of a vehicle in these positions and the center line is also assumend to be normally distributed.

For both measures, we only regard the closest estimated lane with the same semantic driving direction (incoming or outgoing) for the evaluation of each measurement.
We assume the remaining associations to vanish and thus be neglectable in favor of a faster computation time. We also found, that modeling the distribution of the number of trajectories per lane as a multinomial distribution improves the results.

The transition probabilities between the sampling steps $P(I \rightarrow I')$ and vice-versa can be inferred from the probabilities and distributions shown in Algorithm \autoref{al:topo}.
E.g. the probability of changing the angle of an arm by \ang{2} can be formulated as
\begin{equation}
P(\alpha'=\ang{32} | \alpha=\ang{30}) = P_\text{Rot} \cdot P_{\mathcal{U}[\ang{-6},\ang{6}]}(\Delta \alpha = \ang{2}).
\end{equation}
$P_\text{Rot}$ describes the probability of choosing the rotation step\footnote{here: $P_\text{Rot} = 0.3$}.

Finally, the accepted models converge to the maximum of the posterior distribution of the intersection.
With this estimate, the most probable intersection model, given the measured trajectories is determined.
The resulting model, represents the intersection already very precisely, but provides no information on the connectivity of lanes and the lane courses.

\section{Lane Course Refinement}
\label{sec:model}

Using the coarse intersection parameters and the trajectories (cf. \autoref{fig:topo_traj}), an initial estimate on the connectivity of the lane stubs can be made by connecting all stubs passed by the same vehicle (cf. \autoref{fig:split}).
When estimating a connection between the lane stubs, we get an initial lane model, that reduces the refinement of the center lines to a least-squares problem.
When modeled as 1D cubic B-splines we implicitly enforce the lanes to be drivable corridors and the residuum can be determined based on the distance between the splines and the trajectories.

\subsection{Lane Model}
We define each intersection for the lane course estimation as depicted in \autoref{fig:lane_traj} as a set of full lanes $l \in L$.
Each $l = (e, o, Z_l)$ is constructed from an entry lane stub $e \in E$ and an exit lane stub $o \in O$ (stubs are visualized in \autoref{fig:split}).
Each full lane is represented by a cubic B-spline.

Each measurement trajectory is defined as $z \in Z$ which has nearest neighbor assignment $\theta_E: z \rightarrow e$ and $\theta_O: z \rightarrow o$, respectively, to the closest exit $o$ and entry $e$.

\subsection{Initialize Lanes Based on Estimated Parameters}
\label{subsec:estimation}

The estimated intersection parameters of Section~\ref{sec:preproc} are used for initializing the lane course estimation.
The associations $\theta_E$ and $\theta_O$ have been calculated that describes, which entry and exit is closest to a trajectory.
This association can be used to assign each entry lane stub a set of exit lane stubs that are connected by at least on trajectory (cf. \autoref{fig:split}).
We use the association to initialize a lane $l$ with two straight lane stubs ($e$ and $o$) and a straight connection between them.

The angle $\alpha$ of an arm describes the angle of the lane and with the gap width $g$ in combination with the number of lanes $E_a$ resp. $O_a$ and the lane width $w$ an initial lane can be calculated. A resulting initial lane model is depicted in \autoref{fig:split}.

The corridor connecting two associated lane stubs is initialized straight. Since the estimated entry and exit lane stubs of the parameter estimation only have an angle $\alpha$ the initial model can only assume straight arms that will be refined.

\subsection{Uniform Cubic B-Splines}
For the lane refinement, we represent each center line by an 1D cubic B-spline with equidistantly placed knots \cite{deboor_splines_1978}. A cubic B-spline is a piece-wise polynomial function %
defined in variable $x$.

The B-spline curve is defined as a linear combination of $n+1$ control points $c_i$ and B-spline basis functions $N_i,k(x)$ with
\begin{equation}
f_l(x) = \sum_{i=0}^n c_i N_{i,k}(x).
\label{eq:spline}
\end{equation}
Thus, for each part of the spline, $k$ control points influence the positioning and by changing these control points the spline curve can be fitted to any function of interest.
In addition, the resulting spline model is implicitly smooth up to the $(k-2)^\text{th}$ derivative, which is sufficient for lane modeling with $k=4$ \cite{Catmull_spline_1974,Busch_DiscreteReversibleJump_2019}.

\begin{figure}[tb]
	\centering
	\resizebox{0.8\linewidth}{!}{

    \begin{tikzpicture}[
      rotate=180,
      scale=0.6,
      arrow/.style={thin,<->,shorten >=1pt,shorten <=1pt,>=stealth},
      larrow/.style={thick,<-,>=stealth},
      rarrow/.style={thick,->,>=stealth}
    ]
\foreach \y in {-1,0}{
	\draw [thin] plot [smooth] coordinates { (-1, \y)(1,\y) (3,\y) (5,\y) (7,\y)};
}

\foreach \sign in {1}{
	\foreach \x in {0.5,1.5,2.5}{
		\draw [thin] plot [smooth] coordinates {
			  (\sign*\x,\sign*4)
		      (\sign*\x+\sign*0.1*1*1,\sign*4-\sign*1.88*1)
		      (\sign*\x+\sign*0.1*2*2,\sign*4-\sign*1.88*2)
		      (\sign*\x+\sign*0.1*3*3,\sign*4-\sign*1.88*3)
		};
	}
}

\draw [rarrow,fzi-happygreen, dotted, thick] plot [smooth] coordinates { (7,-0.5) (4.5,-0.45) (2.65,0.15) (2.1,2) (2,2.5) (2,4)};
\node at (7.2,-0.5) {\scriptsize A};

\draw [larrow,fzi-yellow, dotted, thick] plot [smooth] coordinates {
	(0.5+1.5+1*0.1*1*1,1*4-1*1.88*1)
	(0.5+1.5+1*0.1*2*2,1*4-1*1.88*2)
	(0.5+1.5+1*0.1*3*3,1*4-1*1.88*3)
};
\draw [larrow,fzi-red, dotted, thick] plot [smooth] coordinates {
	(-0.5+1.5,1*4)
	(-0.5+1.5+1*0.1*1*1,1*4-1*1.88*1)
	(-0.5+1.5+1*0.1*2*2,1*4-1*1.88*2)
	(-0.5+1.5+1*0.1*3*3,1*4-1*1.88*3)
};
\node at (2,-2) {\scriptsize B};
\node at (3,-2) {\scriptsize C};

\draw [larrow, thick, fzi-blue] (1.1,2.5)  -- (2,2.5);
\node at (1.5,2.8) {\tiny $d_\perp$};

\draw [thin] plot [smooth] coordinates { (6.5,0) (4.5,0.05) (3,0.5) (2.59,2) (2.5,4)};
\draw [thin] plot [smooth] coordinates {(6.5,-1) (4.5,-0.95) (2.35,-0.15) (1.6,2) (1.5,4)};

\draw [larrow,fzi-happygreen, thick] (2+3,3.1) -- (2+4.5,1.7) ;
\draw [larrow,fzi-happygreen, thick] (5.1,0.29) -- (2+4.5,1.7) ;
\node[color=fzi-happygreen] at (4.9,3.1) {\tiny x};
\node[color=fzi-happygreen] at (5,0.29) {\tiny y};

\draw [larrow,fzi-red, thick] (6.1,3.585) -- (2+4.5,1.7) ;
\draw [larrow,fzi-red, thick] (4.55,1.24) -- (2+4.5,1.7) ;
\node[color=fzi-red] at (6.1,3.685) {\tiny x};
\node[color=fzi-red] at (4.45,1.24) {\tiny y};

\draw [larrow,thick] (5.5,1.7) -- (2+4.5,1.7) ;
\draw [larrow,thick] (6.5,0.7) -- (2+4.5,1.7) ;
\node at (5.4,1.7) {\tiny x};
\node at (6.5,0.6) {\tiny y};

\draw [rarrow,thick,fzi-happygreen] (4.5,-0.45) -- (4.15,-0.8) ;
\draw [rarrow,thick,fzi-happygreen] (2.65,0.15) -- (2.3,-0.2) ;
\draw [rarrow,thick,fzi-happygreen] (2.1,2) -- (1.75,1.65) ;

\draw [rarrow,thick,fzi-red] (-0.5+1.5+1*0.1*1*1,1*4-1*1.88*1)  -- (-0.5+1.5+1*0.1*1*1-0.48,1*4-1*1.88*1-0.11) ;
\draw [rarrow,thick,fzi-red] (-0.5+1.5+1*0.1*2*2,1*4-1*1.88*2)  -- (-0.5+1.5+1*0.1*2*2-0.48,1*4-1*1.88*2-0.11) ;
\draw [rarrow,thick,fzi-red] (-0.5+1.5+1*0.1*3*3,1*4-1*1.88*3)  -- (-0.5+1.5+1*0.1*3*3-0.48,1*4-1*1.88*3-0.11) ;

\node at (6,-0.25) {\tiny $f_A$};
\node at (1.35,-0.75) {\tiny $f_B$};

    \end{tikzpicture}
 	}
	\caption{Each lane spline (green and red) is defined in a separate coordinate system. The x-axis is aligned with the average direction of all assigned trajectories. Each spline is defined as $f_l: x \rightarrow y$ and thus the optimization only affects the visualized moving direction (small arrows). For the residuals of the optimization, the distance $d_\perp$ as well as the distance between estimated center line and the trajectories is used. Best viewed in color.}
	\label{fig:lane-spline}
\end{figure}
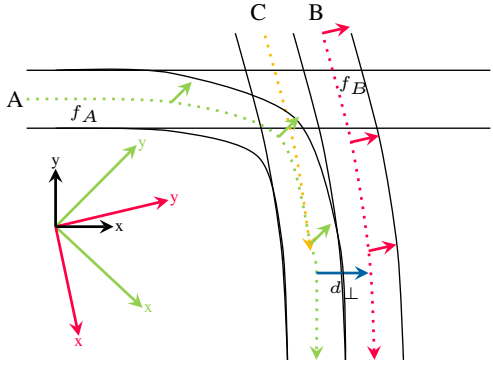

For easing the later fitting of the lanes, we define the spline models in rotated coordinate systems as depicted in \autoref{fig:lane-spline}.
We take the set of all assigned trajectories and set the x-axis of the lane-specific coordinate system (depicted in green and red) to the average direction of all assigned trajectories. The origin is set to the origin of the fixed world coordinate system (depicted in black).

For representing a lane, we use an uniform cubic 1D B-spline. We define the spline as $f_l: x \rightarrow y$ and thus can only adjust the y-coordinate of the lanes (see \autoref{fig:lane-spline}). However, with the rotated coordinate system modifying only one dimension of the problem is sufficient for fitting the lanes to the trajectories while maintaining low complexity of the optimization problem.

\subsection{Least Squares Formulation}
\label{sec:least-squares}

We estimate the lane course and thus implicitly the intersection by constructing a least squares problem using the previously defined lane splines. As stated above the parameters of the cost functor are the 1D control points~$c$\footnote{Here, we use $20$ control points for each lane.}. Using the trajectories $Z_l$, we can optimize the spline parameters.

Given a the point $p = (x,y)$ of a trajectory $z$, we can project the point into the coordinate system of the assigned spline $l$. Given its x-coordinate $p_l^x$ we can calculate the distance along the y-axis to the closest spline point $f_l(p_l^x)$ given with \autoref{eq:spline}. Thus, for each trajectory point we construct the residual as
\begin{equation}
e_1 = (p_{l}^y - f_l(p_l^x))^2.
\end{equation}

After $10$ iterations, we add a second residual to the problem that specifically takes into account that neighboring lanes should have the same border points.
We found it to be beneficial for the quality of the results to first optimize only the lane course and later on regard the bounds.

For a equidistantly sampled set of evaluation points $a$ on a lane spline $A$, we calculate the distance to all lanes $B$, that were estimated as neighbor by the parameter estimation (Section~\ref{sec:preproc}).

We project the evaluation point $a_A$ into the coordinate system of spline $B$. Given the x-coordinate $a_B^x$ and \autoref{eq:spline} we get the point on B with $f_B(a_B^x)$.

Given that, the spline distance $d_\perp$ in point $a_A$ is defined as the dot product of the distance between the two points $a_B^y$ and $f_B(a_B^x)$ and the normal $n_a$ of the spline $A$ in point $a_A$
\begin{equation}
d_\perp = n_a \cdot (a_B^y - f_B(a_B^x)).
\end{equation}

Since we create an initial lane for all combinations of traveled exits and entries (see Section \ref{subsec:estimation}) a single entry (or exit) might have multiple overlapping lanes, that have different destinations (or origins). Lane $A$ and $C$ in \autoref{fig:lane-spline} share an exit but have different entries. Thus, we construct a cost functor $e_2$, that has two minima,
so lanes either fully overlap and thus have a distance $d_\perp = 0$ or be adjacent with $d_\perp = w$.
This way, we can have multiple splines representing the same entry, but diverge into multiple exits and vice-versa with

\begin{equation}
e_2 = ((d_\perp - 0) \cdot (d_\perp - w))^2.
\end{equation}
For the residual, we only regard points, where the distance $d_\perp$ is less than a threshold $\delta > w$.

In a postprocessing step, we combine all estimated lane courses into lanelets~\cite{Poggenhans_Lanelet2highdefinitionmap_2018} and merge lane estimates, that converged onto the same center line, into a single lanelet. This way we end up with a consistent, map-like representation.

\section{Evaluation}
\label{sec:experiments}

\begin{figure*}[tb]
	\begin{center}
		\begin{subfigure}{0.29\linewidth}
			\centering
			\includegraphics[width=1\linewidth]{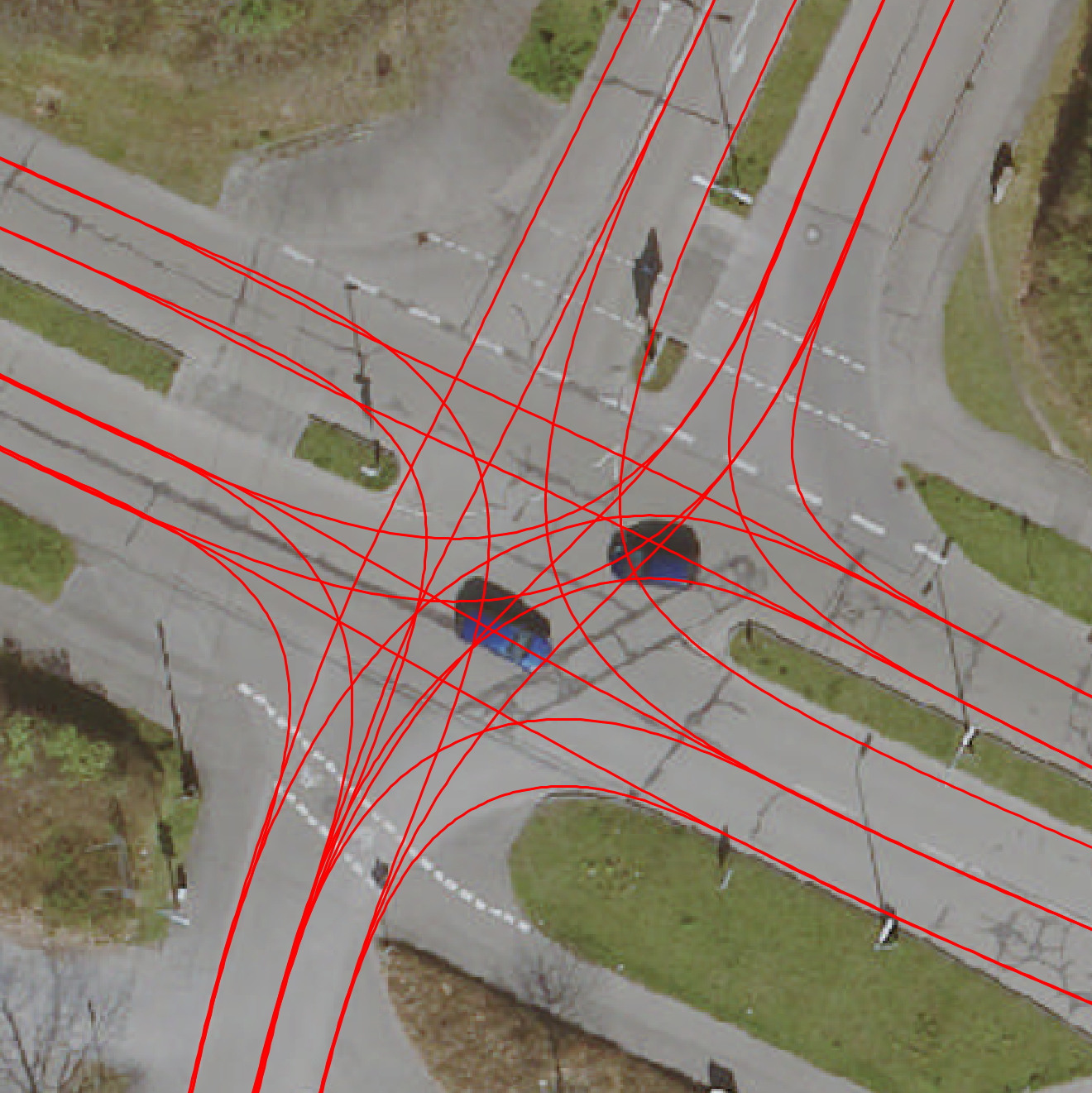}
		\end{subfigure}
		\vspace{2px}
		\begin{subfigure}{0.29\linewidth}
			\centering
			\includegraphics[width=1\linewidth]{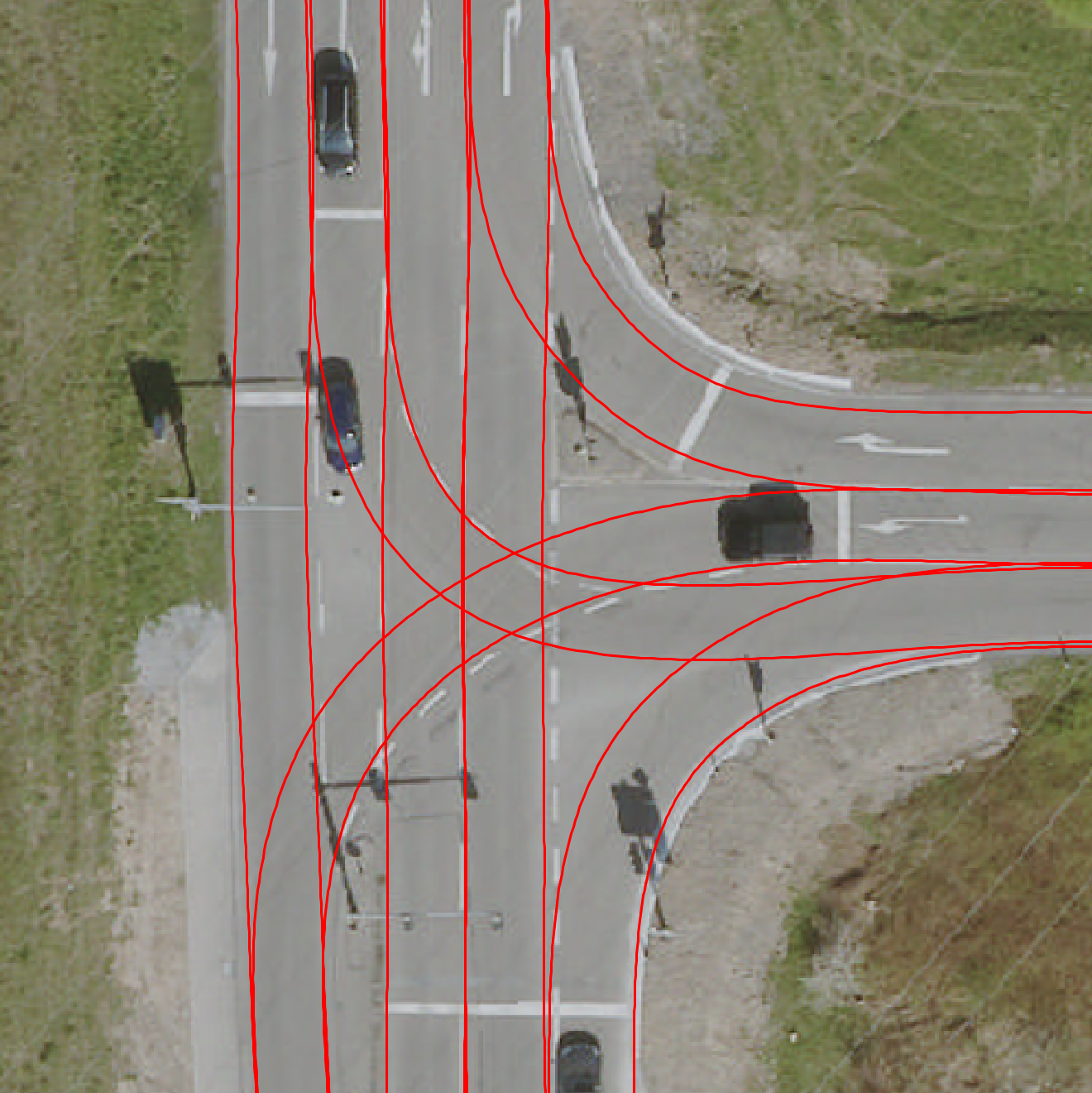}
		\end{subfigure}
		\vspace{2px}
		\begin{subfigure}{0.29\linewidth}
			\centering
			\includegraphics[width=1\linewidth]{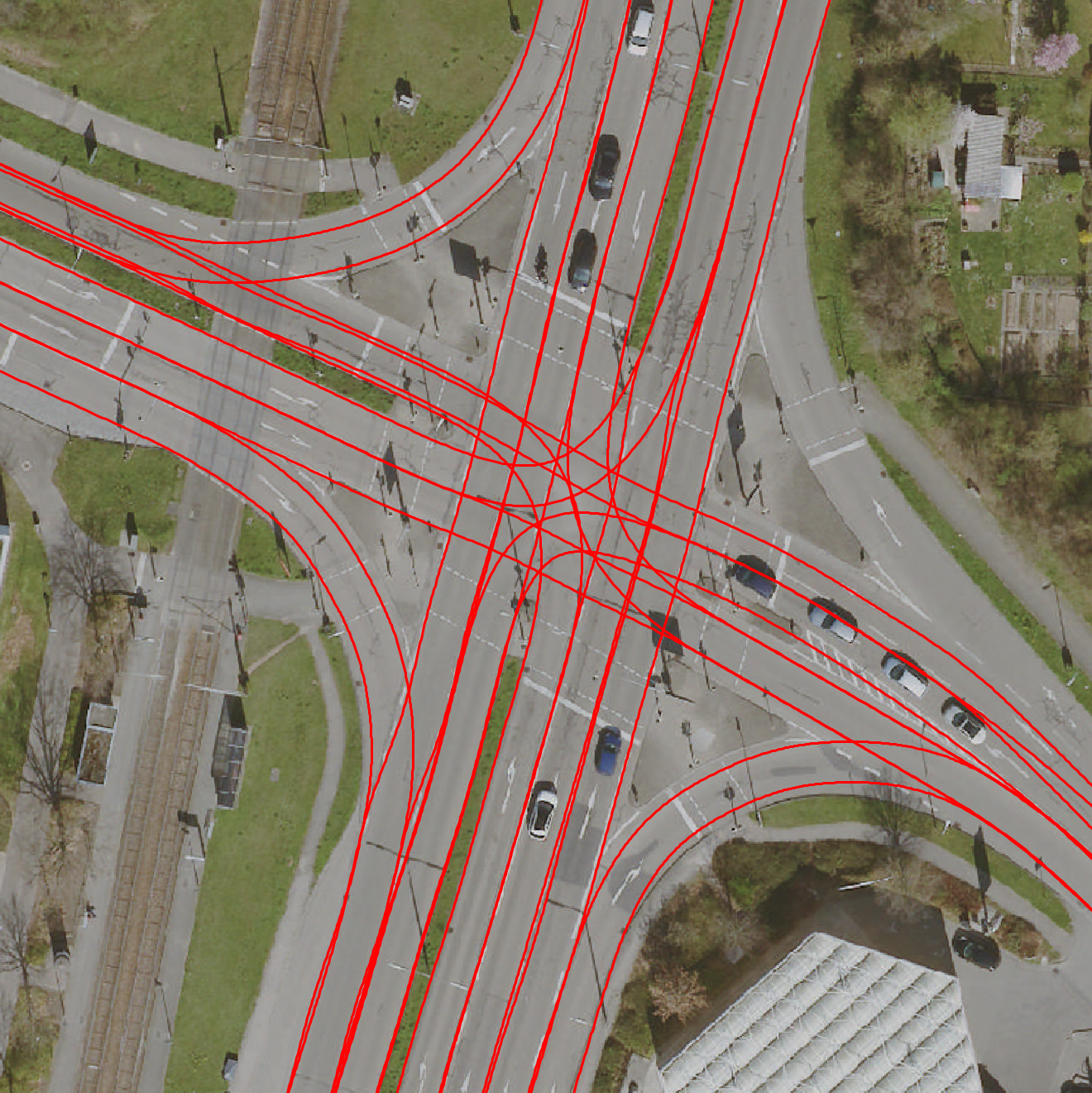}
		\end{subfigure}
		\begin{subfigure}{0.29\linewidth}
			\centering
			\includegraphics[width=1\linewidth]{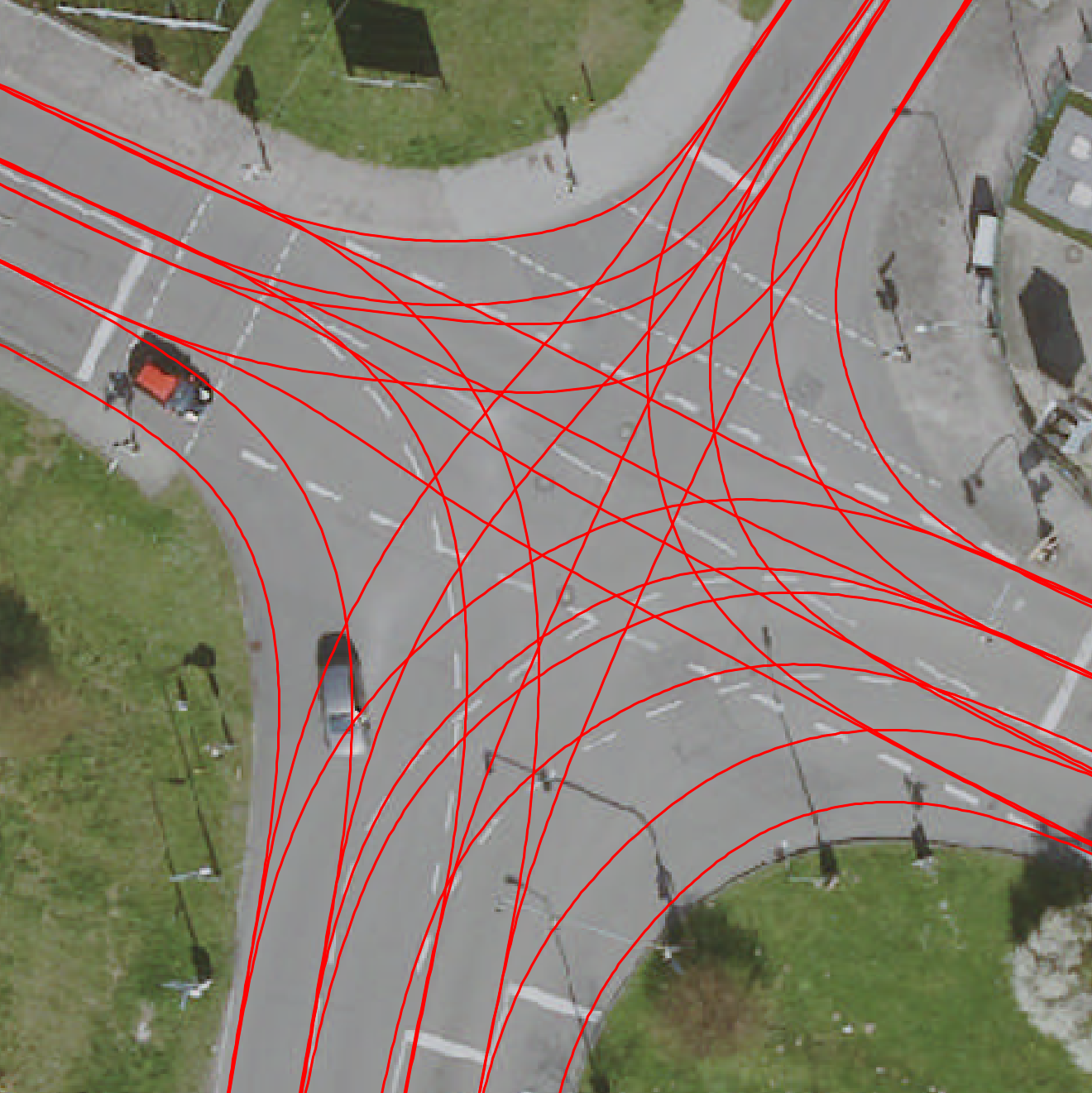}
		\end{subfigure}
		\begin{subfigure}{0.29\linewidth}
			\centering
			\includegraphics[width=1\linewidth]{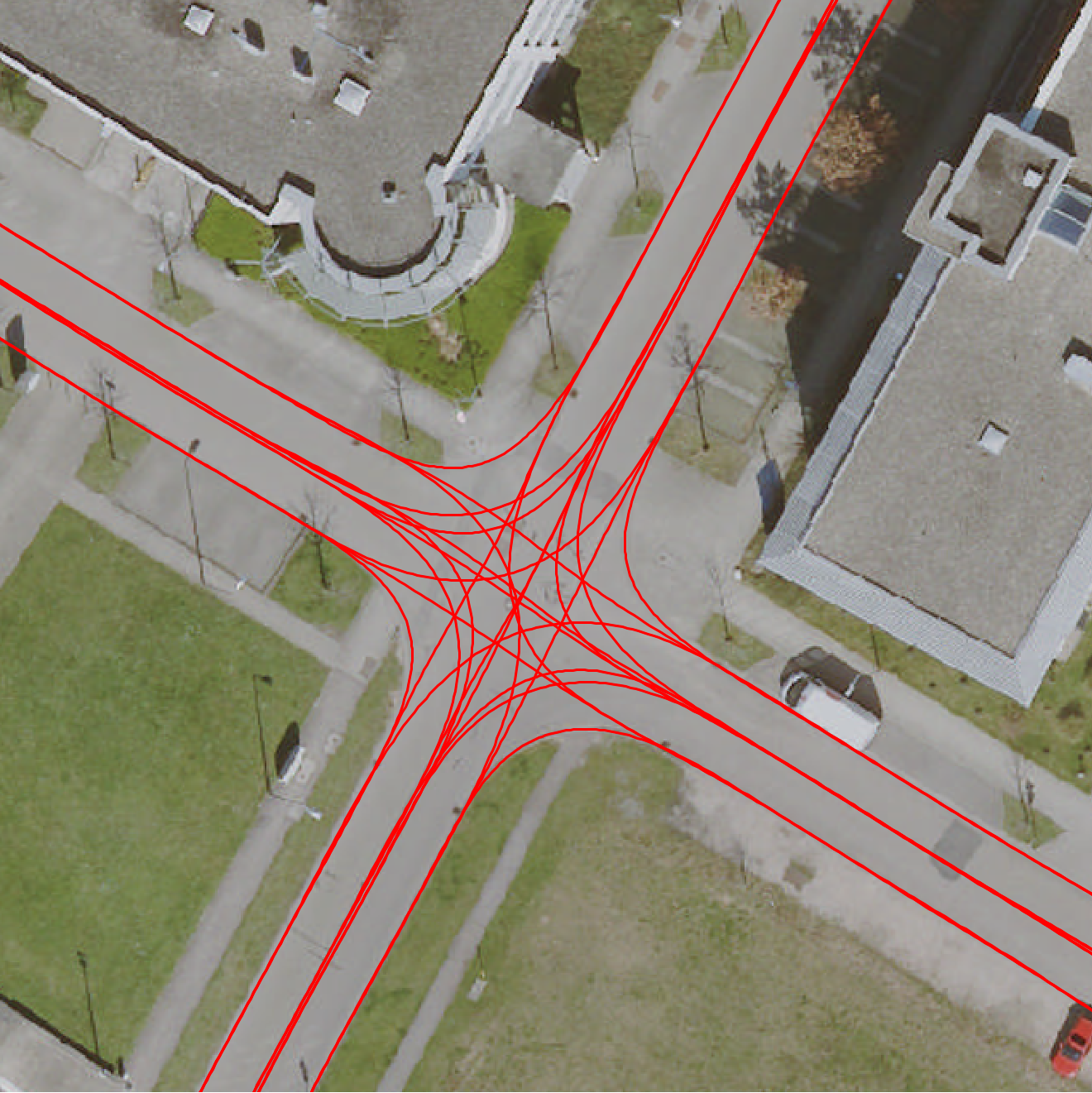}
		\end{subfigure}
		\begin{subfigure}{0.29\linewidth}
			\centering
			\includegraphics[width=1\linewidth]{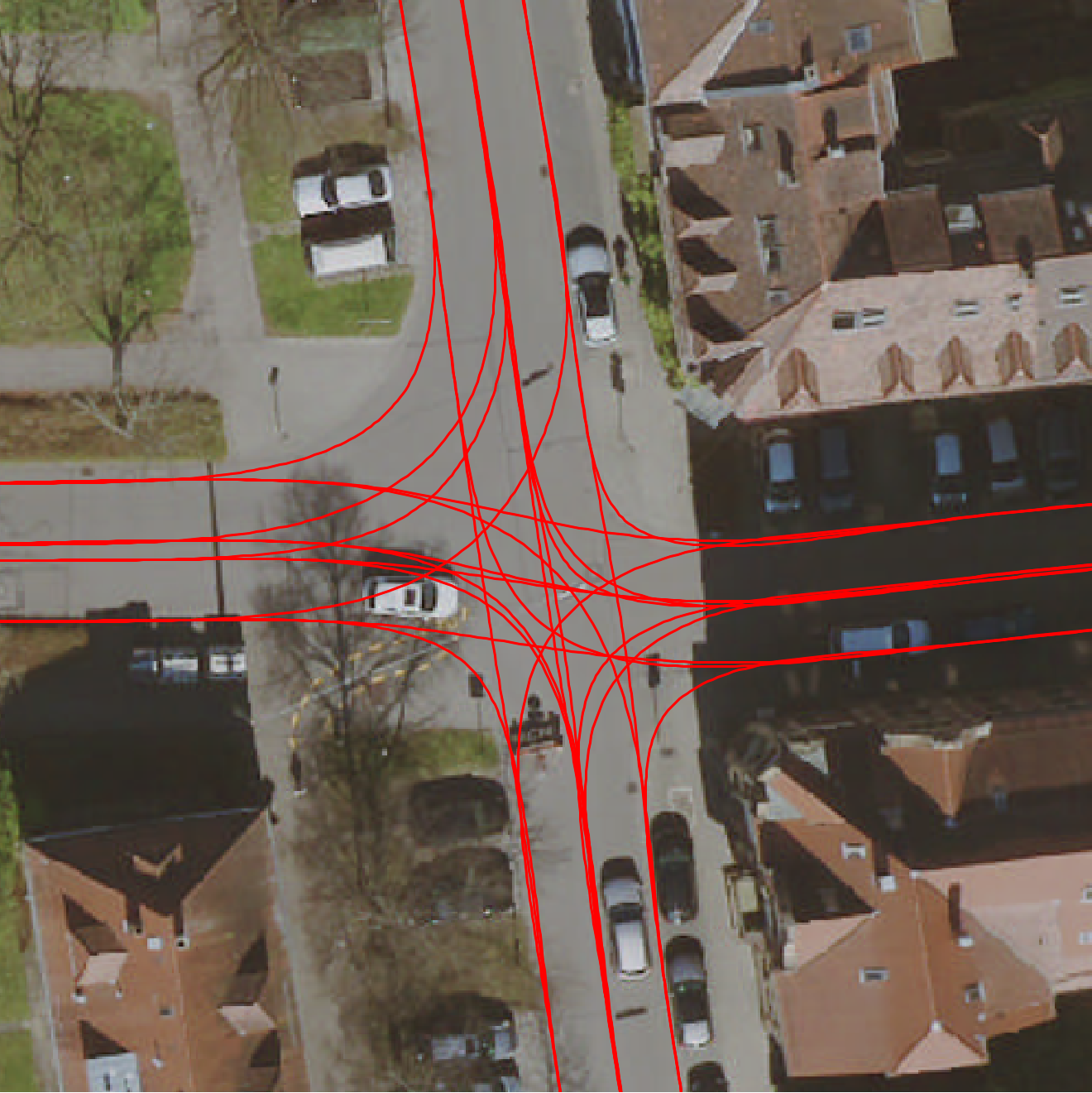}
		\end{subfigure}
	\end{center}
	\caption{Example results of real-world intersections. As we do not estimate the width of a lane, the border visualization can mislead, but the centerlines are estimated with deviations of less than \SI{10}{\centi\meter} on average. Best viewed in color. Aerial images: City of Karlsruhe, \texttt{www.karlsruhe.de}, \texttt{dl-de/by-2-0}} %
	\label{fig:osm_results}
\end{figure*}

The presented approach has been evaluated on 1000 simulated intersections and 14 real-world intersection geometries from Karlsruhe, Germany. For all these intersections we simulated trajectories traveling along each lane. Each lane generated randomly between three and five trajectories. The individual measurement points of the trajectories were equidistantly initialized every \SI{1}{\meter} and distorted by white Gaussian noise $\Delta z_i \gets \mathcal{N}^2(\SI{0}{\meter}, \SI{1}{\meter})$.

The coarse parameter estimation has not been quantitatively evaluated as it did not change significantly compared to \cite{Meyer_AnytimeLaneLevelIntersection_2019}.
For the evaluation of the lane estimation we measure the average orthogonal distance between the estimated lane center line and the center line of the ground truth lanes.

As described in \cite{Meyer_AnytimeLaneLevelIntersection_2019} we can freely weigh the precision against the execution time of the parameter estimation, neglecting convergence quality. Thus, we base our evaluation of the lane estimation on two settings for the parameter estimation. The presented results here are initialized with either the results after \si{5000} or \si{10000} sampling steps in the parameter estimation. The calculation of these samples takes \SI{49}{\milli\second} and \SI{94}{\milli\second}, respectively.

With the estimated parameter model, we can refine the lane courses within \SI{47}{\milli\second} independent of the quality of the parameter estimation. With the \si{5000}-sample parameter model, we achieve an average center line deviation of only \SI{7}{\centi\meter}. If we generate \si{10000} samples in the parameter estimation, we can further improve the result of the lane estimation by \SI{1}{\centi\meter}.
\cite{Meyer_AnytimeLaneLevelIntersection_2019} achieved an average deviation of \SI{14}{\centi\meter} after more than \SI{100}{\milli\second}. Thus, we could halve both the error and the execution time, here.

For real-world geometries we can show, that our approach is able to estimate the lanes with an average deviation of \SI{10}{\centi\meter} after \SI{34}{\milli\second}. Example results for the real-world geometries are depicted in \autoref{fig:osm_results}. We added aerial images for visualization, but only used the geometries for generation of the trajectories and evaluation.

\section{Discussion and Future Work}

We showed an approach that is able to estimate the lane course of intersections up to an average error of only a few centimeters based on trajectories of other traffic participants. Further we could achieve faster execution times of less than \SI{50}{\milli\second} compared to a fully \ac{MCMC} approach by leveraging a least-squares formulation.

Using the trajectories as measurement, we robustify the estimation for heavy traffic situations where visual cues like markings and curbs might be occluded.
Additionally, the calculation of trajectories is ideally based on multiple sensors or at least a sensor like lidar or radar with a far field of view.
As discussed in Section~\ref{sec:related_work}, an estimation purely based on camera information is not sufficient due to distortion induced imprecision.

Our approach is evaluated with simulated data. This is due to the fact, that no dataset exists that contains precise lane information in intersections combined with sensor data from a vehicle.
In the future, we would like to evaluate the approach further on data taken by a measurement vehicle.

Since the model is highly flexible, we can easily add new cues on the road and lane layout. Each measurement has to be represented in a probabilistic manner for the \ac{MCMC} and one has to built a suitable residual for the refinement.

Thus, including markings or curbstones, will be an easy task. When including those or measurements related to the lane width or border, the model can easily be extended to a width estimation both in the parameter estimation for a lane as a whole as well as in the lane refinement for each section of the spline individually.

We believe that this builds a solid foundation for further work in the direction of lane-level intersection estimation for the purpose of mapless driving, map verification and map updates, all in a real-time feasible manner while driving.

\bibliographystyle{ieeetr}
\bibliography{root}
\end{document}